\documentclass[10pt, a4paper]{article}

\usepackage[nolineno,notextfont,nomathfont,notodo,noapacite]{mnn_styles}
\usepackage{lrec-coling2024} 

\usepackage{latexsym}
\usepackage{subcaption}
\usepackage{xurl}

\usepackage[utf8]{inputenc}
\usepackage{fdsymbol}
\usepackage{inconsolata}

\usepackage{multirow}
\usepackage{relsize}

\usepackage{tgpagella} 

\newcommand{\elu}{\fun{elu}}
\newcommand{\gelu}{\fun{gelu}}

\newcommand{\modelRobertaPlus}{\modelname{R\textsubscript{26}}}
\newcommand{\modelXLMRPlus}{\modelname{XLM-R\textsubscript{26}}}

\hypersetup{
    pdfsubject={cs.CL},
    pdfkeywords={Layer Fusion, Transformers, Named Entity Recognition, Few-shot Learning},
}

\name{Muhammad ElNokrashy$^{\ast \mu}$, Badr AlKhamissi$^{\ast \dagger \beta }$, Mona Diab$^{\dagger \omega}$}

\address{
    $^\mu$Microsoft Egypt,
        $^\beta$EPFL,
        $^\omega$CMU \\
    $^\mu$muelnokr@microsoft.com,
        $^\beta$badr.alkhamissi@epfl.ch,
        $^\omega$mdiab@andrew.cmu.edu\\
}

\title{
    Depth-Wise Attention (DWAtt): \\
    A Layer Fusion Method for Data-Efficient Classification
}

\abstract{
    Language Models pretrained on large textual data have been shown to encode different types of knowledge simultaneously.
    Traditionally, only the features from the last layer are used when adapting to new tasks or data.
    We put forward that, when using or finetuning deep pretrained models, intermediate layer features that may be relevant to the downstream task are buried too deep to be used efficiently in terms of needed samples or steps.
    To test this, we propose a new layer fusion method: Depth-Wise Attention ({DWAtt}), to help re-surface signals from non-final layers. We compare {DWAtt} to a basic concatenation-based layer fusion method ({Concat}), and compare both to a deeper model baseline---all kept within a similar parameter budget.
    Our findings show that {DWAtt} and {Concat} are more step- and sample-efficient than the baseline, especially in the few-shot setting. {DWAtt} outperforms {Concat} on larger data sizes. 
    On \dataset{CoNLL-03} NER, layer fusion shows $3.68-9.73\%$ F1 gain at different few-shot sizes.
    The layer fusion models presented significantly outperform the baseline in various training scenarios with different data sizes, architectures, and training constraints.
}

\begin{document}

\maketitleabstract

\blfootnote{\hspace{-0.7em}$^\ast$ Equal contribution.}
\blfootnote{\hspace{-0.7em}$^\beta$ Started independently, finished at Meta AI residency.}
\blfootnote{\hspace{-0.7em}$^\dagger$ Work done while at Meta AI.}
\blfootnote{\hspace{-0.7em}$^\diamond$ Source code available at \href{https://github.com/munael/dwatt-depth_wise_attention-lrec_coling_2024}{github.com/munael/dwatt-depth\_wise\_attention-lrec\_coling\_2024}}

\section{Introduction}

The Transformer architecture \citep{vaswani17attention} and variants \citep{Fan2020AccessingHR} have become a mainstream, reliable choice for a wide range of Natural Language Processing (\newterm{NLP}) tasks, such as sentence classification, token labeling, retrieval, and question answering \citep{wang2018glue, wang2019superglue}.
This is in part due to architectural properties that enable enhanced parallelization and better modeling of long-range dependencies.
Several models have since surfaced, setting new records on different benchmarks \citep{devlin-etal-2019-bert, lewis-etal-2020-bart, gpt3}.
In the vanilla architecture, a stack of Transformer blocks are applied sequentially to refine the representation of an input sequence, which is then fed to a task-specific module, such as a classifier head.

Recent works have shown that hidden representations from intermediate layers may benefit downstream tasks \citep{wallat_bertnesia_2021}.
Other works have tested the fusion of hidden representations in tasks such as sequence-to-sequence machine translation \citep{shen2018dense,liu2020layer,liu2021understanding}.
See also speech modeling works \citep{Li2018Layer}.

In this work, we propose a method to combine the hidden representations of encoder layers to more easily utilize the full model.
Moreover, we further investigate a simpler alternative as a compelling baseline.
This work expands on \textit{Vertical Attention}, introduced in \citet{AlKhamissi_Gabr_ElNokrashy_Essam_2021}, by providing a more detailed specification and an extensive evaluation and analysis.

\begin{figure}
    \centering
    \includegraphics[width=1\linewidth]{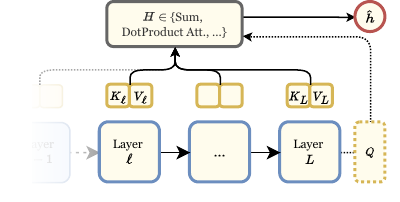}
    \caption{Basic architecture: The Mixer $H$ can be a Sum of Affine Transformations, a Dot-Product Attention module, etc. Different variants may define $K_i$, $V_i$, and $Q$ (key, value, and query transforms) and utilize them differently.
    }
    \label{fig:diagram_1}
\end{figure}

\section{Motivation \& Hypothesis}
A common goal when designing deeper networks is enhancing their ability to represent deeper chains of abstraction in different data domains and training objectives.
As the model is tuned for a specific task, seemingly \emph{unneeded} information is ignored. More general patterns that could benefit different downstream tasks become less likely to pass through intermediate layers to the final representation.

Some base models (like Large Language Models) posses properties which may implicitly mitigate this effect---by leveraging their large parameter capacity and increased width.
The additional parameters enable memorizing more subtleties or idiosyncrasies of the training data, while the increased model width allows for a less compact final representation, which may let through patterns not \emph{directly} used by the general language modeling objective of choice.
Some tasks may need such low-level knowledge across a large example space \citep{liu2021understanding}.
A similar problem is the strength and clarity of gradients into earlier intermediate layers. Methods to alleviate this include skip connections \citep{resnet}, and alternatives to full back-propagation \citep{DFA}.

We propose an add-on module that can augment any pretrained deep sequence model by combining the representations of its intermediate layers to adapt better to novel tasks.
To investigate our proposed modules, we experiment with the following tasks:
\captiona{} Named Entity Recognition (\newterm{NER}) in the few-shot setting on the \dataset{CoNLL-03} and \dataset{WikiAnn} datasets, and
\captionb{} Masked Language Modeling (\newterm{MLM}) on a small text dataset (\dataset{WikiText-2}). Two settings are considered: Finetuning (\newterm{FT}) and Feature Extraction (\newterm{FE}).

\paragraph{Adapting to New Tasks.}
For many pretraining projects, adaptability to novel language tasks is a crucial performance signal. it allows one to exploit general language understanding from self-supervised tasks like \oldterm{MLM} on large datasets.

Useful signals include: \captiona{} Performance versus a model trained on task-specific data only, \captionb{} time to convergence, and \captionc{} size of the needed task data to achieve the desired performance.
 
One could finetune the whole model (possibly infeasible), or use it as a feature extractor and train only a task-specific module. A commonly researched alternative is finding or introducing a subset of parameters specifically for finetuning \citep{adapters,BenZaken2022BitFitSP}.
We test our proposed method and an alternative in different adaptation settings with multiple n-shot training sizes, max training steps, and tiers of availability of language data during pretraining.
%


\begin{figure}
    \centering
    \includegraphics[width=0.9\linewidth]{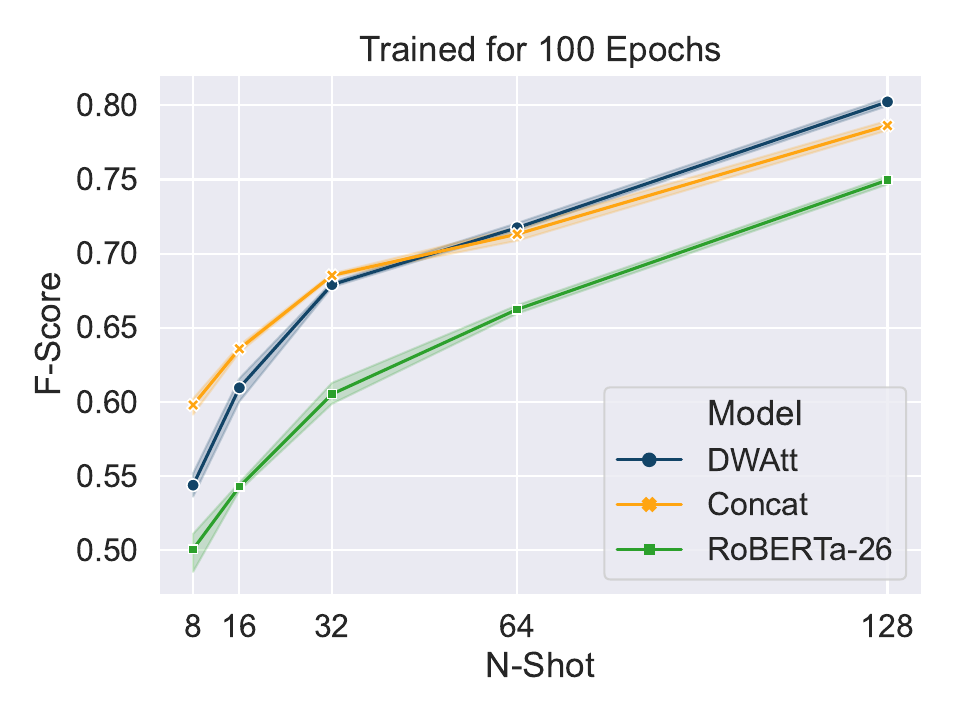}
    \caption{%
        F1-Score on the \dataset{CoNLL-03} dev set after training for $100$ epochs on few-shot training datasets with \textit{N-Shot} samples per class sampled uniformly from the full training set.
        Both \modelname{DWAtt} and \modelname{Concat} improve on the additional layers baseline \modelRobertaPlus{}. At $128$ samples per class, \modelname{DWAtt} improves on both \modelname{Concat} and the enhanced baseline by $1.6\%$ and $5.28\%$ absolute, respectively.
    }
    \label{fig:conll-100}
\end{figure}


\section{Tasks, Datasets, and Raised Questions}

On many common benchmarks, state-of-the-art performance is often near saturation. We consider some useful synthesizable variants of the benchmarks that test particular training settings or aspects of performance. Our experiments aim to test the following aspects: \textit{adaptability} (finetuning (FT) versus feature extraction (FE)), \textit{sample efficiency} (few-shot training), \textit{training step efficiency} (time to convergence), and \textit{effect of model depth}.

\subsection{Few-Shot Adaptation on NER}
\begin{table}[h]
\centering
\begin{tabular}{@{}lcccc@{}}
\toprule
    \textbf{Dataset}
    & \textbf{Labels}
    & \textbf{Resource}
    & \textbf{Train}
    & \textbf{Dev}  \\ \midrule
\dataset{CoNLL-03}      
& 4
        & High      & 14k    & 3250 \\ \midrule
\multirow{3}{*}{\dataset{WikiAnn}}
& \multirow{3}{*}{3}
          & High      & 20k    & 10k  \\
        & & Mid       & 1k-5k & 1k   \\
        & & Low       & 100    & 100  \\ \bottomrule
\end{tabular}
\caption{Train and Dev subset sizes in sentence count. For \dataset{WikiAnn}, we list the average size of a language in the corresponding resource tier, determined by train size.}
\label{tab:ner-data}
\end{table}

NER, a moderately complex task, evaluates performance at a token-level, and reasonably can benefit from intermediate features. See section \ref{ssec:eval-selection}.

\paragraph{\dataset{CoNLL-2003}} The \dataset{CoNLL-03} dataset \citep{tjong-kim-sang-de-meulder-2003-conll} provides a mainstream NER benchmark in the English language. For few-shot trials, we sample training points uniformly at random from the full training set.

\paragraph{\dataset{WikiAnn}} The \dataset{WikiAnn} dataset \citep{rahimi-etal-2019-wikiann} is a multilingual NER benchmark which we use to test the effect of resource availability in the pretraining phase. For few-shot experiments, we sample a fixed training set of size $N \teq 100$ and train for each language separately. We also compare performance on the English subset between RoBERTa and the multilingual XLM-RoBERTa.


\subsection{Masked Language Modeling}
\label{sec:mlm}

The \dataset{WikiText-2} dataset \citep{Merity2017PointerSM} is a subset of the English Wikipedia developed for use in long-term dependency language modeling. It contains the first $2$ million words of a dump of the English Wikipedia at the time of creation.
We report the perplexity on the dev and test sets. We hypothesize that retaining better pretraining performance points to increased reliability. See \Secref{ssec:eval-selection}.


\section{Models}
\label{sec:method}
Let $L$ denote a deep network's layer stack.
Then $\lparam{H}$\footnotemark\ is a learned function that mixes the layers' intermediate representations $\{\vz_\In \mid \In \in \abs{L}\}$, and ${\vh}$ is the final representation.

\footnotetext{In this work, symbols in \lparam{red} have learned parameters.}
\begin{equation}
    {\vh}
    = \lparam{H}~(
        \cdots,
        \big\{ (\In, \vz_\In) \mid \In \in \abs{L} \big\}
    )
\end{equation}
The $H$ module is a function of the layer indices $\In$ and the corresponding representation vectors $\vz_\In$, but can take other signals, like $\vz_L\mapsto\vq$ for query.
See diagram in \Figref{fig:diagram_1}.
We consider the following layer fusion models.


\paragraph{Layer Concatenation.}
A sum of linear transforms.
Note that this is equivalent to concatenating all $\{\vz_\In\}$ then transforming the concatenation into the model width $\dimD{z}$.
\begin{equation}
    {\vh} = {\textstyle\sum_\In}\ \lparam{\mW_\In}(\vz_\In)
\end{equation}

\paragraph{Depth-Wise Attention.}
\modelname{DWAtt} uses dot-product attention with
keys $\vk_\In$, values $\vv_\In$, and query $\vq$.
\begin{align}
    \vk_\In &= \lparam{\fun{PE}}(\In) \\
    \vv_\In &=
        \lparam{\fun{LN}_\In}\left(\lparam{f^V_\In}(\vz_\In)\right)
    \\
    \vq      &=
        1 + \elu\left(
            \vz_L + \lparam{f^Q}(\vz_L)
        \right)
\end{align}
Where $\fun{PE}(\In)$ is a learned positional embedding vector for layers $\{\In\}$.
$\fun{LN}$ is LayerNorm \citep{Ba2016LayerNorm}.
$f^Q$, $f^V_\In$ are MLPs with a bottleneck at $1/2$ the model width $\dimD{z}$.
$\fun{elu}$ is from \citet{elu}.
\begin{equation}
    f(\vz) = \lparam{\mW}\cdot\lparam{\func{LN}}(\gelu(\lparam{\mU} \vz))
\end{equation}
It is then comparable to a single linear layer of size $\dimD{z}\times\dimD{z}$. See \Secref{app:design} for details.
Note: the scoring step in $\fun{Attend}$ reduces to a single, vector-by-static matrix multiplication as the keys $\{\vk_\In\}$ need no input.
Assume a time step $t$, then let $\mK=\{\vk_\In\}$ and $\mV=\{\vv_\In\}$ be the matrix forms of the keys and values for $\In\in\abs{L}$. Then:
\begin{align}
    \func{Score}(\vq, \mK) &=
        \fun{softmax}_\In
        \left(
            \vq \cdot \mK^\top
        \right)
    , \\
    \func{Attend}(\cdots) &=
            \fun{Score}\pa{\vq, \mK}
            \cdot \mV
    , \\
    {\vh} &=
        \vz_L + \fun{Attend}\big(
            \vq, \{\vk_\In\}, \{\vv_\In\}
        \big).
\end{align}

\subsection{Baselines}
\begin{table}[ht!]
\centering
\begin{tabular}{@{}lcrr@{}}
\toprule
\textbf{Name}    & \textbf{New Layer} & \textbf{+ $O(\cdot)$} & \textbf{+ \#}
    \\ \midrule
\textbf{Base}    & {-}              & -
                                    & - \\
\textbf{Base\textsubscript{+n}}
                 & {$n\times$ Base} & $7n\ (d + d^2)$
                                    & $25.19$M \\
                 \midrule
\textbf{Concat}  & {Affine}         & $Ld^2$
                                    & $25.18$M \\
                 \midrule
\textbf{DWAtt}   & {DWAtt}          & $Ld^2 + d^2 + Ld$
                                    & $26.38$M \\
                 \bottomrule
\end{tabular}
\caption{%
The \textbf{size and count of parameters of new layers} in different configurations. Counts are of models building on RoBERTa\textsubscript{24}.
\modelname{Base\textsubscript{+n}} (deeper baseline) is chosen to be close in size to \modelname{DWAtt} and \modelname{Concat}.
For example, for base models with 12 layers, only n=1 layers are added; n=2 for 24 layer models. $L$ refers to the depth of the \emph{base} model, while $d$ refers to the model width.
}
\label{tab:arch-comp}
\end{table}%
\paragraph{Base Model.} The base model (\oldterm{RoBERTa} or \oldterm{XLM-RoBERTa} in \textsc{base} or \textsc{large} sizes) is used as-is. The task module is changed where needed. In \hterm{FE} only the task module (or vocabulary module in \oldterm{MLM}) is trained.

\paragraph{Extra Transformer Layers.} On top of the base model, we add $2$ more Transformer layers before the classification head. In \hterm{FE} mode, we train only the added layers. We refer to this model as \newterm{\modelRobertaPlus{}} (\oldterm{RoBERTa}) or \newterm{\modelXLMRPlus{}} (\oldterm{XLM-RoBERTa}).


\section{Experimentation Setup}
\subsection{Base Models}
Monolingual experiments using \dataset{CoNLL-03}, the English subset of \dataset{WikiAnn}, and \dataset{WikiText-2} all build on \text{RoBERTa} (\newterm{R}) \citep{Liu2019RoBERTa} via the pretrained \texttt{roberta-large} model. Multilingual experiments using \dataset{WikiAnn} build on \text{XLM-RoBERTa} (\newterm{XLM-R}) via the pretrained \texttt{xlm-roberta-large} model. Experiments using other variants are explicitly specified. Pretrained models are accessed via the HuggingFace package \citep{Wolf2019HuggingFace}.

\subsection{Training} 
\label{sec:training}
In figures where the epoch count is reported: Each chart shows an experiment with the indicated $\dimN{epochs}$ max epochs, which starts LR at max then decays it linearly to zero.

Few-shot experiments sample $NC$ points uniformly at random from the full training set. Sampling is not stratified, so $N=8$ Shot for $C=4$ classes means $32$ points in total sampled without replacement.


\subsection{Evaluation}
NER experiments report Micro-Average F1 using seqeval \citep{seqeval}.
Where applicable, experiments are given 5 trials whose average and confidence interval are reported by \code{seaborn} \citep{seaborn}. For easier reading, we report scores that lie in $[0,1]$ as percentages $[0,100]\%$.


\section{Results and Analysis}
\label{sec:results}

\subsection{\dataset{CoNLL}: Few-Shot Adaptation}
\begin{figure*}[!ht]
    \centering
    \includegraphics[width=\linewidth]{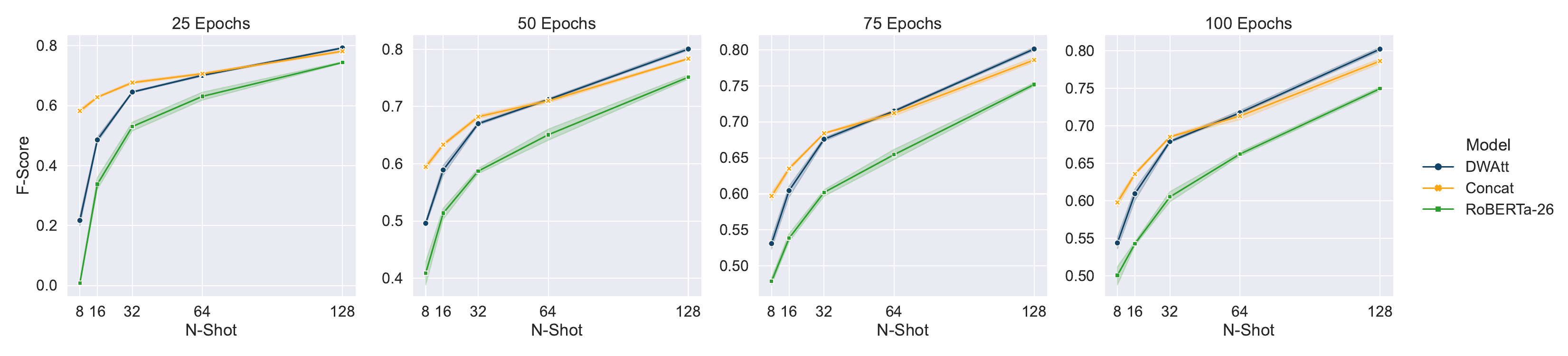}
    \caption{\textbf{F1-Score on the \dataset{CoNLL-03} devset.} All pretrained weights are frozen (\hterm{FE}). In each chart from left to right, training is constrained to $25$, $50$, $75$, and $100$ max epochs. For each \textit{N-Shot} experiment, $NC$ samples ($C=4$ classes) are randomly selected and trained on for the entire experiment. The scores are averaged across $5$ trials with random initialization of weights and data sampling. We report the best observed dev score from the full training of each experiment and trial.}
    \label{fig:conll-nshots}
\end{figure*}
Micro-F1 is reported at each \textit{N-Shot} in $\{8,16,32,64,128\}$.
Each model in \Figref{fig:conll-nshots} adds the specified module on top of a pretrained \oldterm{RoBERTa\textsubscript{LARGE}}.
\modelname{Concat} (orange) noticeably outperforms \modelRobertaPlus{} at all few-shot settings, while \modelname{DWAtt} (blue) outperforms \modelname{Concat} at the higher data sizes.

\subsection{Step and Sample Efficiency}
\label{sec:efficiency}
\begin{figure}[ht]
    \centering
    \begin{subfigure}[b]{0.5\linewidth}
        \centering
        \includegraphics[width=\linewidth]{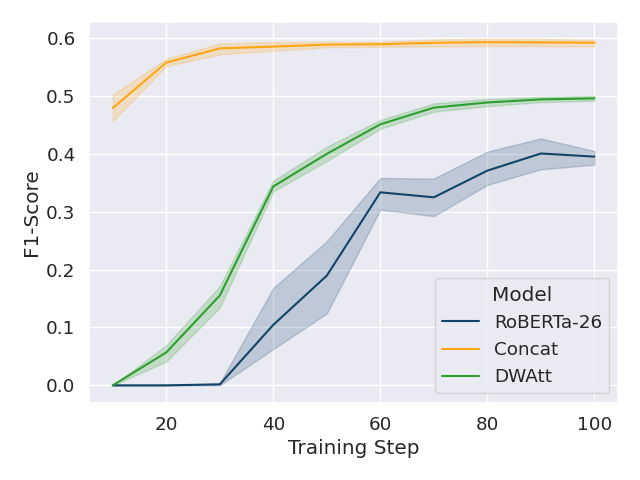}
        \caption{Few-shot at $N=8$.}
    \end{subfigure}%
    \begin{subfigure}[b]{0.5\linewidth}
        \centering
        \includegraphics[width=\linewidth]{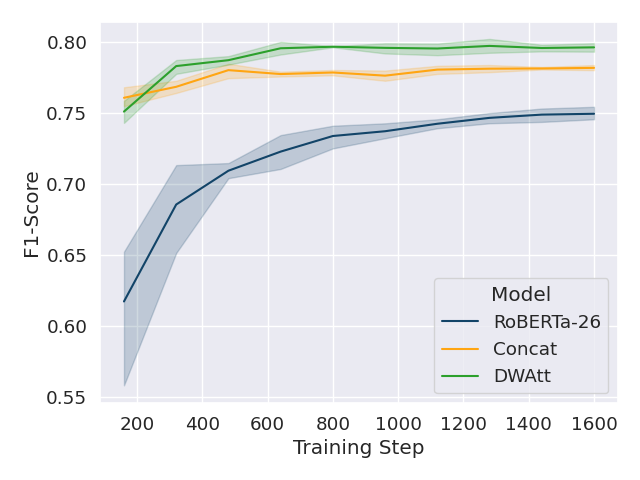}
        \caption{Few-shot at $N=128$.}
    \end{subfigure}
\caption{%
\textbf{Training Behavior.}
Validation F1-Score across steps in \hterm{FE} training on \dataset{CoNLL-03}. \modelname{Concat} generalizes more readily at smaller $N$ compared to \modelname{DWAtt}. At both sizes, layer fusion methods are able to extract more from pretrained models than traditional last-layer fitting.
}
\label{fig:conll-training-f1}
\end{figure}

In \BFigref{fig:conll-nshots}, \modelname{Concat} outperforms noticeably at the lowest end ($N{=}8$ at $\dimN{epochs}{=}25$), improving on \modelRobertaPlus{} by $\mathbf{58\%}$---possibly owing to its simpler connectivity and gradient path.
While \modelname{DWAtt}'s improvement given increased data and training time may be explained by
its being more selective due to the attention module.
In the largest setup ($N{=}128$, $\dimN{epochs}{=}100$; \SFigref{fig:conll-100}),
\modelname{Concat} improves on \modelRobertaPlus{} by $3.68\%$, while \modelname{DWAtt} improves on it by $\mathbf{5.28\%}$.
\modelRobertaPlus{} demonstrates the difficulties of extracting the patterns needed by a downstream task from a feature extractor that has already 
fit its last layer representation ($L{=}24$ for \oldterm{RoBERTa\textsubscript{LARGE}}) for its pretraining task.

In \BFigref{fig:conll-training-f1}, we focus on model validation behavior during training with $\dimN{epochs}{=}50$. At $N{=}8$, only \modelname{Concat} manages to adapt well, and rapidly, converging in $30\%$ of training time. At $N{=}128$, \modelname{DWAtt} has enough data to reach similar performance at a similar pace, and quickly pulls ahead.

\begin{figure*}[!ht]
    \centering
    \includegraphics[width=\linewidth]{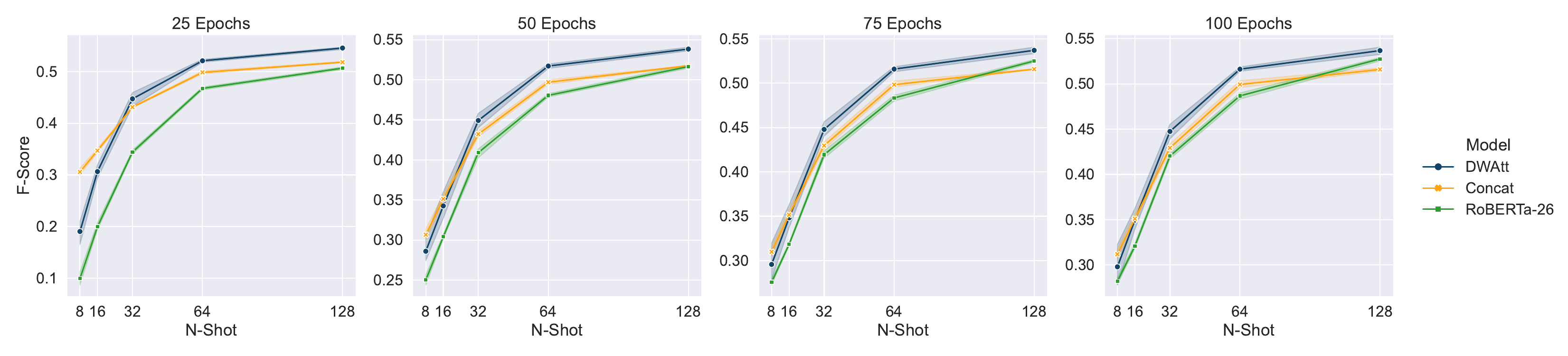}
    \caption{F1-Score on the \dataset{WikiAnn} \emph{English} devset on the \emph{\oldterm{RoBERTa\textsubscript{LARGE}}} base model. The methods start behaving similarly at higher training epochs. \modelname{DWAtt} leads \modelname{Concat} leads \modelRobertaPlus{} at medium shot scenarios.}
    \label{fig:wikiann-en-all}
\end{figure*}

We repeat in \BFigref{fig:wikiann-en-all} the experiments from \SFigref{fig:conll-nshots} but on English only from \dataset{WikiAnn}, on \modelRobertaPlus{}. Yet again \modelname{Concat} performs better with less data, while \modelname{DWAtt} is better at the higher end ($N{=}128$, $\dimN{epochs}{=}100$) by $+2.08\%$ and $+0.92\%$ over \modelname{Concat} and \modelRobertaPlus{}, respectively.

\subsection{Feature Extractor Adaptability}
\begin{figure}[t]
    \centering
    \begin{subfigure}[b]{0.5\linewidth}
        \centering
        \includegraphics[width=\linewidth]{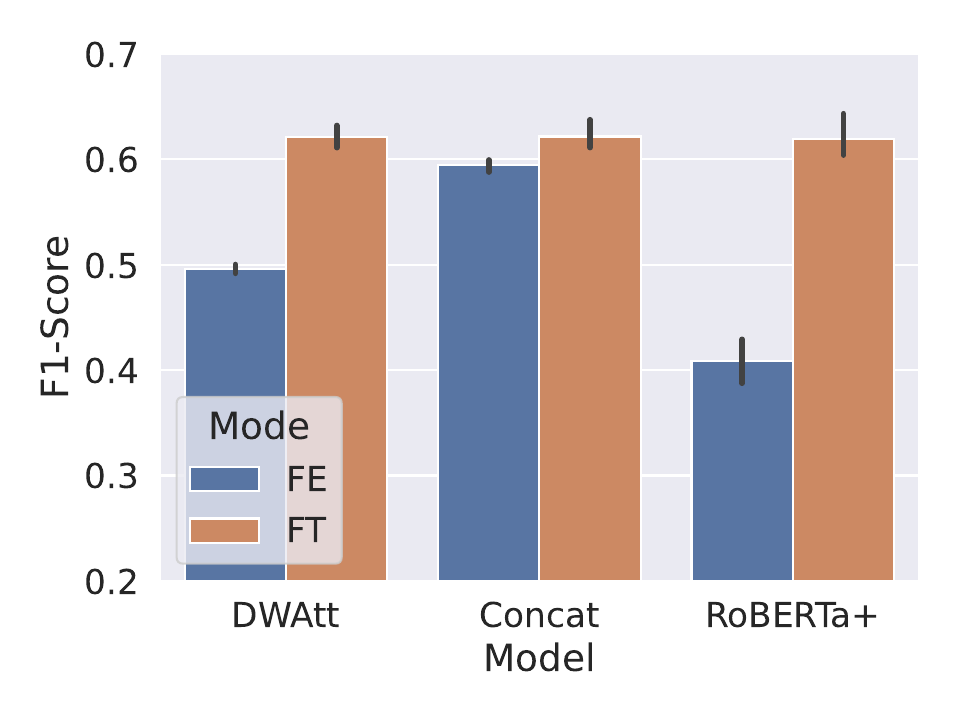}
        \caption{$N=8$}
    \end{subfigure}%
    \begin{subfigure}[b]{0.5\linewidth}
        \centering
        \includegraphics[width=\linewidth]{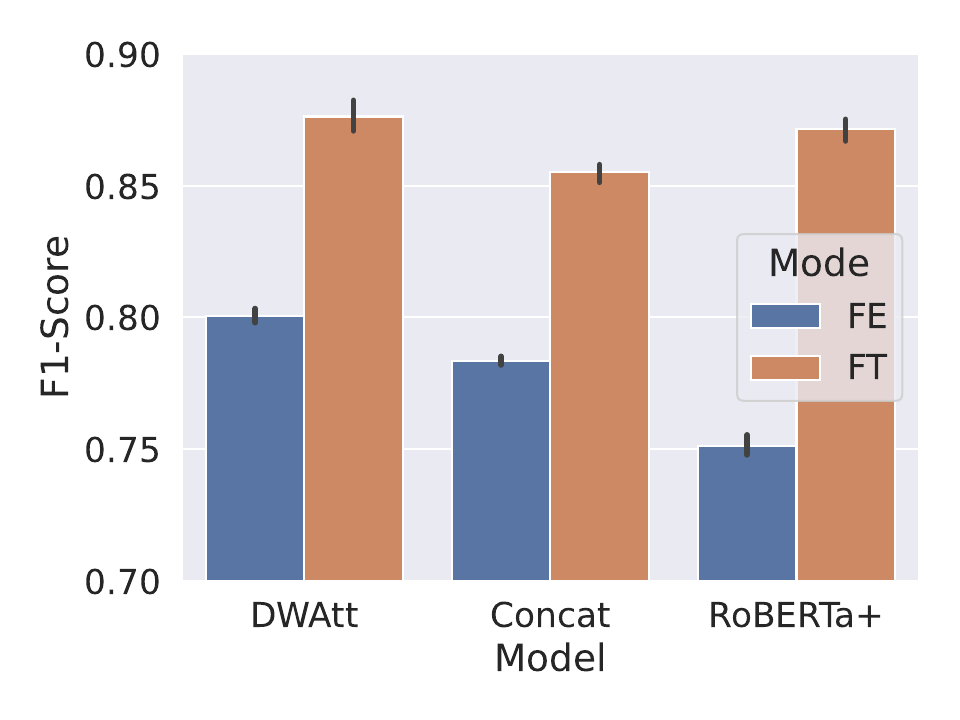}
        \caption{$N=128$}
    \end{subfigure}
\caption{%
\textbf{Feature Extraction (\emph{FE}) versus Finetuning (\emph{FT}).}
Best validation F1-Score on \dataset{CoNLL-03} within $\dimN{epochs}=50$ training epochs. Finetuning consistently outperforms alternatives, but \modelname{Concat} and \modelname{DWAtt} approach its performance even in \hterm{FE} training at lower data sizes.
}
\label{fig:conll-ft-fe}
\end{figure}

\BFigref{fig:conll-ft-fe} shows the effect of finetuning (\hterm{FT}) all model parameters for each of the same configurations used in other \dataset{CoNLL-03} experiments.
At $N=8$, \modelname{Concat} in feature extraction (\hterm{FE}) training already gets most of the improvement observable from full \text{FT} across the board.
At $N=128$ the effect is stronger. While layer fusion methods help close the gap---with \modelname{DWAtt} giving nearly half the improvement of \text{FT} using only \text{FE}, at $+5.28\%$---\text{FT} still manages overshadow all \text{FE} configurations even with only \modelRobertaPlus{}.

\subsection{Scaling by Depth}
\begin{figure}[h]
    \centering
    \begin{subfigure}[b]{0.5\linewidth}
        \centering
        \includegraphics[width=\linewidth]{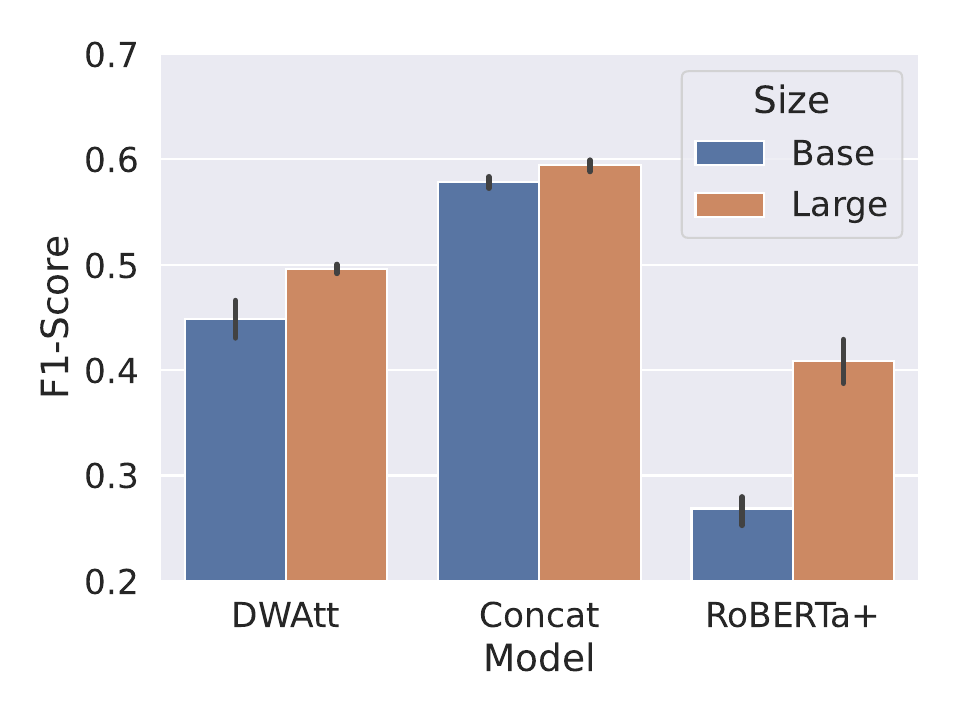}
        \caption{$N=8$}
    \end{subfigure}%
    \begin{subfigure}[b]{0.5\linewidth}
        \centering
        \includegraphics[width=\linewidth]{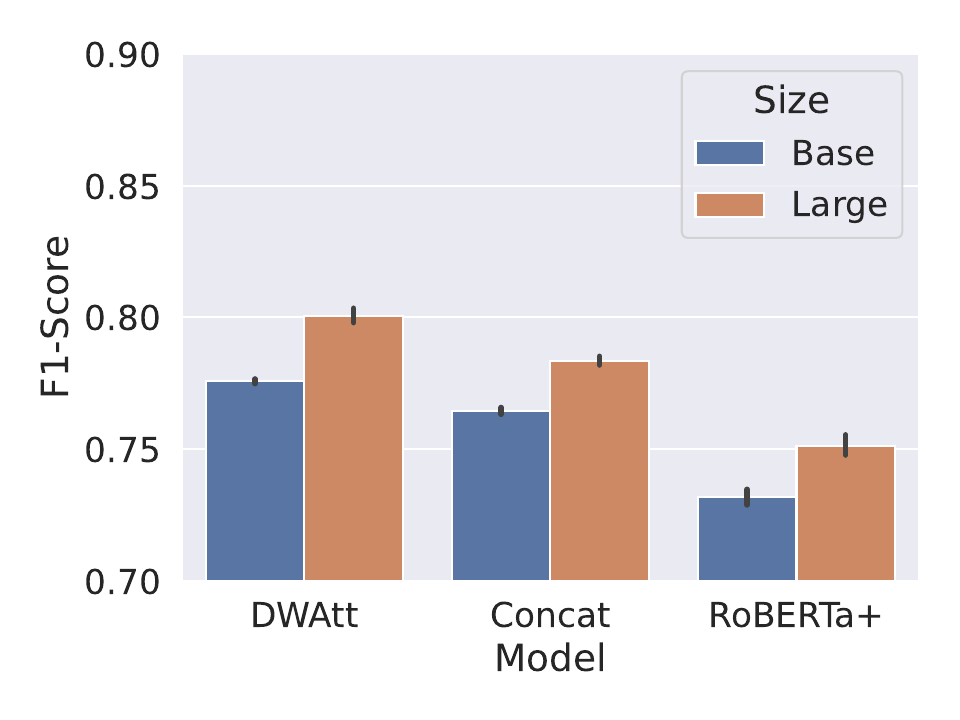}
        \caption{$N=128$}
    \end{subfigure}
\caption{%
\textbf{\textsc{base} versus \textsc{large} Pretrained Models.}
Validation F1-Score on \dataset{CoNLL-03} on the \modelname{DWAtt}, \modelname{Concat}, and enhanced (+layers) configurations of RoBERTa \textsc{base} and \textsc{large}.
\captiona{} At $N\!=\!8$, \modelname{Concat\textsubscript{BASE}} shows a clear lead even on \modelname{RoBERTa\text{+}\textsubscript{BASE}}.
\captionb{} At $N\!=\!128$, \modelname{Concat\textsubscript{BASE}} outperforms \modelname{RoBERTa\text{+}\textsubscript{LARGE}}, while \modelname{DWAtt\textsubscript{LARGE}} outperforms \modelname{Concat\textsubscript{LARGE}}.
\captionc{} This also supports the claim for higher data and training efficiency from \modelname{Concat} and \modelname{DWAtt} in \hterm{FE} training compared to traditional last-layer fitting.
}
\label{fig:conll-base-vs-large}
\end{figure}

In \BFigref{fig:conll-base-vs-large}, we compare the same add-on configurations on the \textsc{base} and \textsc{large} variants of RoBERTa. At $N=8$, \modelname{Concat} is an outlier in its performance gain over the alternatives. In all $N\in\{8,128\}$, both \modelname{Concat} and \modelname{DWAtt} match or beat the performance gain observed from \modelname{RoBERTa\text{+}}\footnotemark{} when changing from \textsc{base} to \textsc{large}.
\footnotetext{\modelname{RoBERTa\text{+}} \textsc{base} is $L{=}12+1$ layers, while \textsc{large} is $L{=}24$+2.}


\begin{figure*}[!ht]
    \centering
    \includegraphics[width=\linewidth]{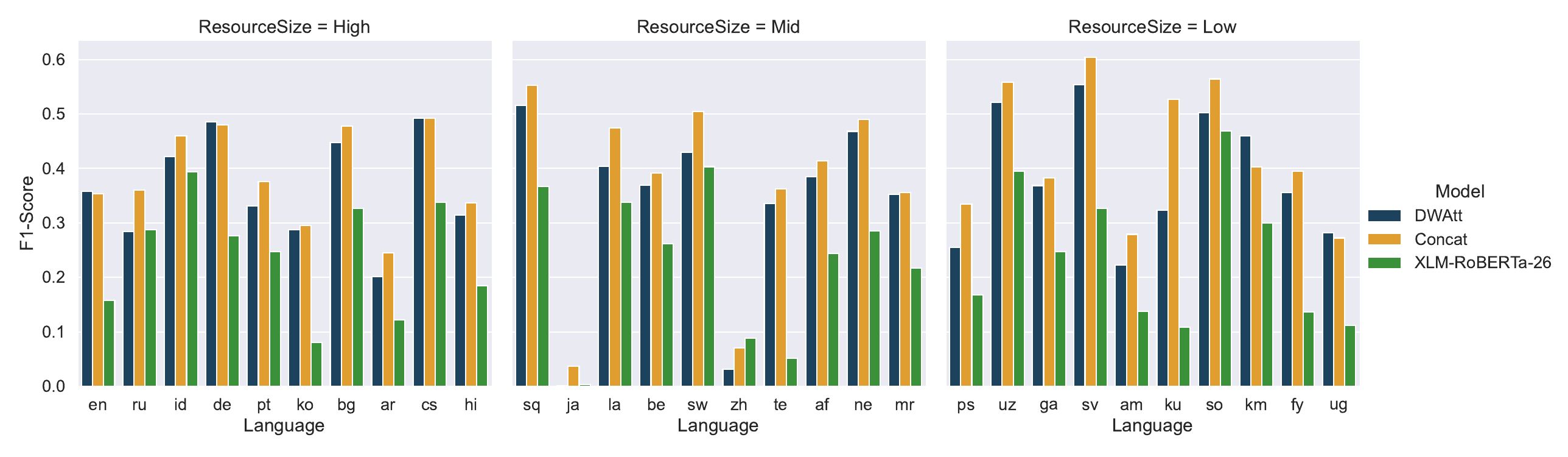}
    \caption{F1-Score on the \dataset{WikiAnn} devset. Each language is trained for $\dimN{epochs}=25$ and tested separately. Each model is initialized from \oldterm{XLM-R\textsubscript{LARGE}} weights, augmented with the specified module (DWAtt, Concat, or additional layers), then trained on exactly $100$ uniformly-sampled training points for each language. The languages are sorted by token count of pretraining data from \citet{xlm-roberta}.}
    \label{fig:wikiann-all}
\end{figure*}
\subsection{\dataset{WikiAnn}: Adapting to High, Mid, and Low Resource Languages}
\begin{figure}[t]
    \centering
    \includegraphics[width=0.9\linewidth]{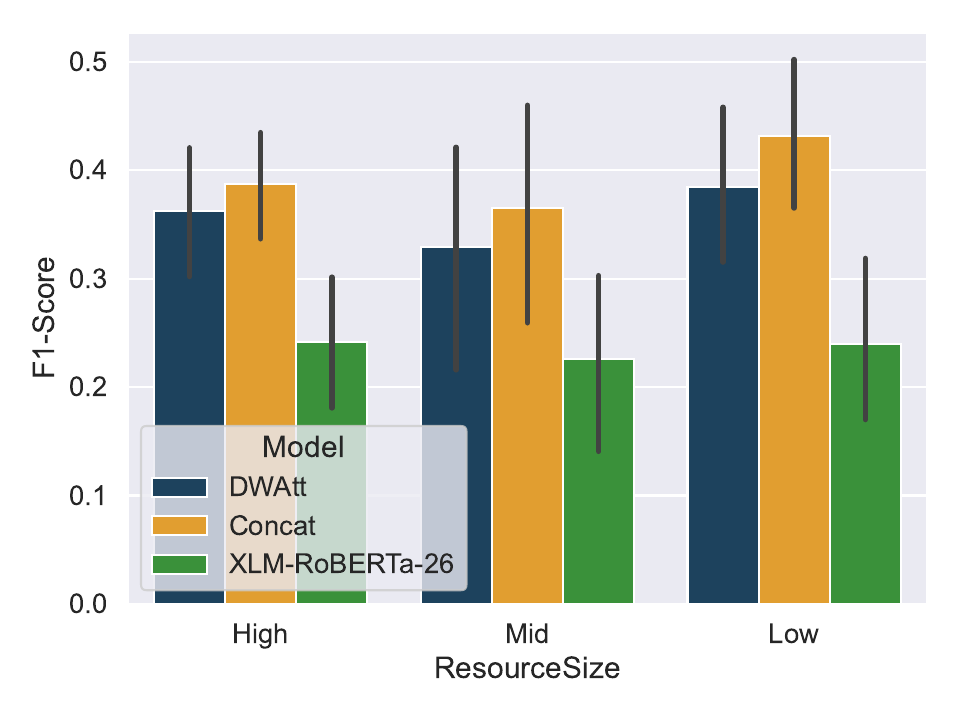}
    \caption{F1-Score on the \dataset{WikiAnn} devset. Languages are grouped into three tiers of resource size. See \Figref{fig:wikiann-all} and \Tableref{tab:ner-data} for details.}
    \label{fig:wikiann-grouped}
\end{figure}

\BFigref{fig:wikiann-grouped} shows average performance grouped by the token count of pretraining data according to \citet{xlm-roberta}. Low-resource languages had $10$--$100$M tokens, Mid-resource languages had $200$--$300$M tokens , while High-resource languages had $2$--$20$G tokens.
While there is a slight lead for layer fusion methods in the Low-resource column, the difference does not translate to a similar lead in Mid-resource scores over High-resource.


Compare the scores in both \textbf{Figures} \textbf{\ref{fig:wikiann-en-all}} \& \textbf{\ref{fig:wikiann-all}}. Starting from $\dimN{epochs}{=}25$ and only $N{=}32$ samples, performance using \oldterm{RoBERTa\textsubscript{LARGE}} as the base model beats the corresponding run on \oldterm{XLM-R\textsubscript{LARGE}} at $\dimN{epochs}{=}25$ with $100$ samples.


\subsection{Masked Language Modeling}

\begin{table}[h]
\centering
\begin{tabular}{@{}cc@{}}
\toprule
\textbf{Model} & \textbf{Test Perplexity} \\ \midrule
RoBERTa+       & 2.98                     \\
Concat         & 3.44                     \\
DWAtt          & 2.91                     \\
Base           & 2.90                     \\ \bottomrule
\end{tabular}
\label{tab:lm-scores}
\caption{MLM perplexity on \dataset{WikiText-2} test set.}
\end{table}

We report the test perplexity of the checkpoint with the best validation loss from FT training. \modelname{DWAtt} performs closer to the baseline \modelname{Base} than \modelname{Concat} and \modelRobertaPlus{}.

\section{Related Work}

\paragraph{Adaptors.} Adaptors are layers introduced into a pretrained model layer stack and finetuned on a specific downstream task, to avoid changing the (larger) model's weights. \citep{adapters}, among others.
\paragraph{Layer Aggregation (static weights).}
\citet{bapna2018training} defines, for each decoder layer, a trainable softmax-normalized vector of weights to get the weighted sum of the encoder intermediates.
\citet{liu2021understanding} provides one static, learned vector for each intermediate layer's representation which acts as element-wise scaling.
Both works are similar to \modelname{Concat} in applying a static, learned mixing transform to all layers.
For each layer: The first provides a scalar weight,
the second provides a vector for element-wise multiplication,
while \modelname{Concat} applies a linear transform.

In \citet{shen2018dense}, the \newterm{DenseNMT} is an encoder-decoder NMT architecture densely connected in the style of \newterm{DenseNet} \citep{huang2017densenet}. Each encoder layer takes a concatenation of all previous layer representations. Similarly for decoder layers. See also \citet{wang2019learning}. These methods compare structurally to \modelname{Concat}, which is applied only once on the full layer stack.

\citet{wang2018multi} generates the weight for an encoder layer via an MLP on the layer's representation, irrespective of the rest of the model: $\evw_{i}=f^a(\vz_i)$ then ${\vh} = \sum_{i}(\evw_i \vz_i)$.

\paragraph{Dynamic Layer Mixing.}
The following methods use signals from the input to change the transformation itself dynamically.
See also \modelname{DWAtt} which uses a Dot-Product Attention module for comparison.
\citet{Li2018Layer} applies an LSTM depth-wise on the intermediate vectors of the stack of LSTM cells applied on input sequences, as expected.
We attempted adding an LSTM cell applied depth-wise to the Transformer encoder stack but observed lacking performance. Note that the referenced work utilizes a non-basic cell with peep-hole expressions, and some architecture connectivity that complicates experiments in the Feature Extraction context.
\citet{dou2019dynamic} utilizes a dynamically-weighted routing mechanism to mix transformations of each intermediate representation, then concatenate all such.

\section{Discussion}
\subsection{Notes on Selected Evaluations}
\label{ssec:eval-selection}

\paragraph{NER: A Token-level Downstream Task}
The NER task is selected to provide a more complex performance signal than sentence classification (which was the case in \citealp{AlKhamissi_Gabr_ElNokrashy_Essam_2021}).
See \Secref{ssec:modeling-capacity} where we hypothesize on the cases that would benefit from DWAtt. Briefly, tasks which look at individual tokens deeply and do not need increased spatial abstraction (for example because of increased context length) may be more likely to benefit from methods like DWAtt, especially if full fine-tuning is not available.
Further, prior work such as \citet{wallat_bertnesia_2021} suggests that there can be useful information in intermediate layers, and specifically probes for it by a token-level task.

\paragraph{MLM: Pretraining objective}
Users build on pretrained large models in order to exploit a more general linguistic and logical capability than could be expected from training on a downstream task dataset much smaller than the smallest of unsupervised LM data collections.
We believe that fine-tuning must maintain this more general objective of the base model in order to make the claim that task performance generalizes outside the specific sub-domain of the test set. If fine-tuning results in great downstream results but equally great pretraining objective regression (e.g. MLM), we believe that method should be regarded as less reliable and desirable than one which maintains the pretraining objective scores.

\subsection{Levels of Abstraction}
For non-recurrent deep neural networks, there exists a functional limit to the depth of abstraction obtainable, which is proportional to the depth of the network.
Abstraction here refers to the depth of composed rules that a model may learn to apply on low-level stimulus, such as pixels or tokens. For feed-forward models like an MLP or transformers, this corresponds to the depth in terms of stacked nonlinear layers.
As an example, to handle program-like systematic inputs, depth-recurrent and memory-augmented architecture were utilized in works such as \citet{universal-transformer,neural-turing-machine}.
For traditional Transformer models, like \oldterm{RoBERTa\textsubscript{LARGE}}, we can say that the limit is proportional to the number of layers in the network (e.g. $O(|L|)$; $|L|=24$).

\paragraph{Hiding Within Scale.}
\todo{This was more relevant to LAMA. Is it still useful?}
In practice, this is seldom an obvious problem because large networks would have enough width-wise parameter capacity to directly encode ``intuition'', i.e. shortcuts to knowledge and abstraction.
They may do so by tying the low-level representation of some inputs to intermediate signals for the high-level concepts they tend to manifest.
As an artificial example: A small, shallow model for classifying sentiment may tie a token such as \code{scary} to become a strong signal for negative sentiment, say in a movie review setting. Sensible in the domain of kids movies; but may in fact signify a positive sentiment instead when observed in reviews for horror movies. The key point is that it is an early shortcut, not whether it is accurate.

\paragraph{Intuition as Shortcuts From Raw Input.}
\todo{This was more relevant to LAMA. Is it still useful?}
Thus, earlier layers can encode information at a higher level than $|L|$-steps of abstraction would suggest. This would be needed, and expected, for models of a reasonable depth and width to be able to satisfy the feature extraction needs of the \emph{pretraining} task at the last layer. Should a \emph{downstream} task require a different signal from what the \emph{pretraining} task exposed, it may have to \emph{shift} a large subset of the weights to relearn or resurface that information from the shallower levels.

\paragraph{Proposition.}
The methods we've discussed may enable downstream tasks to query the model for hidden but useful information. For the downstream task to make use of such features, it would likely need to transform them further.
By applying the $2$-layer MLPs $f^V_\In$ on these intermediate features (in \modelname{DWAtt}), and a linear transform in \modelname{Concat}, the task can extract a more useful representation from each level/layer.

\subsection{Modeling Capacity}
\label{ssec:modeling-capacity}
The two models presented and highlighted---\modelname{DWAtt} and \modelname{Concat}---are aggregate views of the features of all intermediate Transformer layers $\{\vz_\In \mid \In \in |L|\}$ (see \Secref{sec:method}).
\modelname{DWAtt}'s added module requires access to the last layer's $\vz_L$ to form the query, while \modelname{Concat} does not.
Neither method makes any use of external signals such as, for example, a task embedding vector.

These three points together present an underlying property of the modeling capacity of \modelname{DWAtt} and \modelname{Concat} when \emph{fully-tuned}:
\emph{The depth-wise layer mixing arrives at a model which is at most as expressive as the underlying sequence-modeler}.
See \Figref{fig:conll-ft-fe} where layer fusion under FE nears but does not exceed FT, while all methods are similar under FT.

\paragraph{Modeling Dimensions.}
\modelname{DWAtt} and \modelname{Concat} operate \emph{depth-wise} over a sequence-modeling model. By that very nature, it may not be the best option when the task at hand requires increased or improved \emph{spatial} abstraction---the ability to learn connections on a spatial axis (e.g. between tokens in sequences in a text Transformer).
Then, adding extra Transformer layers or full finetuning may be better options.

\section{Design Details}
\label{app:design}
\begin{table}[h!]
\centering
\begin{tabular}{@{}clc@{}}
\toprule
\textbf{Model} & \multicolumn{1}{c}{\textbf{Param}} & \textbf{Value} \\
\midrule
    \textbf{All\textsubscript{Large}}
                   & Transformer Layers  & 24             \\
    \textbf{All} & Learning Rate & 1e-5 \\
\midrule
    \textbf{DWAtt} & $\gamma^Q$, query bottleneck   & 0.5  \\
    \textbf{DWAtt} & $\gamma^V$, values bottleneck  & 0.5  \\
    \textbf{DWAtt} & ${\dimD{pos}}$, keys latent    & 24   \\
\bottomrule
\end{tabular}
\caption{Architecture Parameters}
\label{tab:arch-params}
\end{table}

\subsection{Training}
\todo[inline,color=red]{Update for NER.}
All trainings use the AdamW \citep{adamw} optimizer with a linear decay learning rate (LR) schedule. Training on \dataset{WikiText-2} uses a batch size of $8$ samples, and a max LR of $\num{5e-5}$. Training on \dataset{CoNLL-03} and \dataset{WikiAnn} uses a batch size of $16$ and a max LR of $\num{5e-5}$.

\subsection{Layer Index Embedding}

\(\vk^{\symb{pos}}_\In\!\in\R^{\dimD{pos}}\!\!\sim\mathcal{U}(0,1)\) are static positional embedding vectors of the layer index $\In$. $\mW^K$ is a learned affine transform.
\begin{equation}
    \vk_\In = \fun{PE}(\In) = \lparam{\mW^K} \vk^\symb{pos}_\In
\end{equation}
\subsection{MLP Modules}
\label{app:mlp}
For each role $\in \{Q, V\}$ that uses an MLP, one is defined for a layer $\In$ as:
\begin{align}
    f_\In(\vx) &= 
    \lparam{\mW_{\In}}\cdot\func{LN}\left(
        \gelu\left(\lparam{\mU_{\In}}~\vx\right)
    \right)
\end{align}
Where $\mU_{\In}, \mW_{\In}$ are the down and up projections of the bottleneck ($\dimD{z} \mapsto \gamma\dimD{z} \mapsto \dimD{z}$), respectively. This means, for example, that for value transforms one set of weights is assigned to each layer $\In$ with nothing shared. $\gelu$ is the activation function from \citet{gelu}.


\section{Conclusion \& Future Work}
\todo[inline,color=yellow]{@bkhmsi - Review.}
We present \modelname{DWAtt}---a new method of reusing the latent representations of a deep neural network.
We analyze \modelname{DWAtt} and a similar, simpler method---\modelname{Concat}---from multiple aspects of performance and scaling on NER and MLM tasks.
Results suggest similar layer fusion methods can be a strong tool for downstream adaptation. Performance gains from $1\%$ to $6\%$ and as high as $30\%$ can be seen in various experiments for different few-shot sizes and training times, under Finetuning or Feature Extraction, and on base models of different depths. \modelname{DWAtt} and \modelname{Concat} have shown improved performance even in Feature Extraction training against full Finetuning.
We believe this effect may extend to other tasks besides sequence labeling and to other sequence modeling architectures besides the Transformer---and propose such analysis for future work.
We believe \emph{add-on}-style additions to pretrained models, such as adapters and depth-wise mixing, to be a fertile ground for research into low-cost adaptation of large models that has not been saturated yet.

\section{Bibliographical References}\label{sec:reference}
\bibliography{references}
\bibliographystyle{lrec-coling2024-natbib}

\end{document}